\documentclass[11pt,a4paper]{article}
\usepackage{authblk}
\usepackage[hyperref]{emnlp-ijcnlp-2019}
\usepackage{times}
\usepackage{latexsym}
\usepackage{float}
\usepackage{url}
\usepackage{array}
\usepackage{booktabs}
\usepackage{arydshln}
\usepackage{graphicx}
\usepackage{amssymb}
\usepackage{amsmath}
\usepackage{bm}
\usepackage{multirow}
\usepackage[bb=boondox]{mathalfa}
\usepackage{xspace}
\usepackage{color,soul}
\usepackage{subfigure}
\usepackage{enumitem}
\setitemize{noitemsep,topsep=0pt,parsep=0pt,partopsep=0pt}

\newcolumntype{L}[1]{>{\raggedright\let\newline\\\arraybackslash\hspace{0pt}}m{#1}}
\newcolumntype{C}[1]{>{\centering\let\newline\\\arraybackslash\hspace{0pt}}m{#1}}
\newcolumntype{R}[1]{>{\raggedleft\let\newline\\\arraybackslash\hspace{0pt}}m{#1}}

\newcommand{\bsy}[1]{\boldsymbol{#1}}
\newcommand{\mb}[1]{\mathbf{#1}}
\newcommand{\cev}[1]{\reflectbox{\ensuremath{\vec{\reflectbox{\ensuremath{#1}}}}}}
\DeclareMathOperator*{\argmax}{arg\,max}

\DeclareMathOperator*{\AGGREGATE}{Aggregate}
\DeclareMathOperator*{\softmax}{Softmax}

\aclfinalcopy

\title{Learning to Infer Entities, Properties and their Relations from Clinical Conversations}

\author{Nan Du}
\author{Mingqiu Wang}
\author{Linh Tran}
\author{Gang Li}
\author{Izhak Shafran}
\affil{\{dunan, mingqiuwang, tranlm, leebird, izhak\}@google.com}
\affil{Google Inc.}

\date{}

\begin{document}
\maketitle

\begin{abstract}
Recently we proposed the Span Attribute Tagging (SAT) Model~\cite{symPaper:ACL2019} to infer clinical entities (e.g., symptoms) and their properties (e.g., duration). It tackles the challenge of large label space and limited training data using a hierarchical two-stage approach that identifies the span of interest in a tagging step and assigns labels to the span in a classification step.

We extend the SAT model to jointly infer not only entities and their properties but also relations between them. Most relation extraction models restrict inferring relations between tokens within a few neighboring sentences, mainly to avoid high computational complexity. In contrast, our proposed Relation-SAT (R-SAT) model is computationally efficient and can infer relations over the entire conversation, spanning an average duration of 10 minutes. 

We evaluate our model on a corpus of clinical conversations. When the entities are given, the R-SAT outperforms baselines in identifying relations between symptoms and their properties by about 32\% (0.82 vs 0.62 F-score) and by about 50\% (0.60 vs 0.41 F-score) on medications and their properties. On the more difficult task of jointly inferring entities and relations, the R-SAT model achieves a performance of 0.34 and 0.45 for symptoms and medications respectively, which is significantly better than 0.18 and 0.35 for the baseline model. The contributions of different components of the model are quantified using ablation analysis. 
\end{abstract}

\section{Introduction} \label{sec:intro}
The widespread adoption of Electronic Health Records by clinics across United States has placed a disproportionately heavy burden on clinical providers, causing burnouts among them~\cite{HBR2018,Atlantic2018,Arndt:2017}. There has been considerable interest, both in academia and industry, to automate aspects of documentation so that providers can spend more time with their patients. One such approach aims to generate clinical notes directly from the doctor-patient conversations~\cite{PatDavPan18, FinleyEdwards:2018, FinSalSad18}. The success of such an approach hinges on extracting relevant information reliably and accurately from clinical conversations. 

In this paper, we investigate the tasks of jointly inferring entities, specifically, symptoms (Sx), medications (Rx), their properties and relations between them from clinical conversations. These tasks are defined in Section~\ref{sec:task}. The key contributions of the work reported here include: (i) a novel model architecture for jointly inferring entities and their relations, whose parameters are learned using the multi-task learning paradigm (Section~\ref{sec:rsat}), (ii) comprehensive empirical evaluation of our model on a corpus of clinical conversations (Section~\ref{sec:expts}), and (iii) understanding the model performance using ablation study and human error analysis (Section~\ref{sec:analysis}). Since clinical conversations include domain specific knowledge, we also investigate the benefit of augmenting the input feature representation with knowledge graph embedding. Finally, we summarize our conclusions and contributions in Section~\ref{sec:conclusions}.

\section{Task Definitions}
\label{sec:task}
For the purpose of defining the tasks, consider the snippet of a clinical conversation in Table~\ref{table:task}.
\begin{table}[ht]
\centering
{\small
  \begin{tabular}{l}
\textbf{DR:} How often do you have \ul{pain} in your arms? \\
\textbf{PT:} It hurts \ul{every morning}.\\
\textbf{DR:} Are you taking anything for it? \\
\textbf{PT:} I've been taking \ul{Ibuprofen}. \ul{Twice a day}. 
  \end{tabular}
}
 \caption{An example to illustrate entities, properties and their relations. Entities -- (sym/msk/pain: {\it pain}) \& (meds/name: {\it Ibuprofen}); Properties -- (symprop/freq: {\it every morning}) \& (medsprop/freq: {\it twice a day}); Relations: (sym/msk/pain, symprop/freq, {\it every morning}), (Ibuprofen, medsprop/freq, {\it twice a day}).}
 \label{table:task}
\vspace*{-.1in}
\end{table}

\subsection{The Symptom Task (Sx)}
This task consists of extracting the tuples ({\it symType, propType, propContent}). 

The ``pain'' in the example in Table~\ref{table:task} is annotated as {\it symType} ({sym/msk/pain}), where msk stands for musculo-skeletal system. We have pre-defined 186 categories for symptom types, curated by a team of practising physicians and scribes, based on how they appear in clinical notes. We deliberately abstained from the more exhaustive symptom labels such as UMLS and ICD codes~\cite{Bodenreider2004-hv} in favor of this smaller set since our training data is limited.

The properties associated with the symptoms, {\it propType}, fall into four categories: symprop/severity, symprop/duration, symprop/location, and symprop/frequency. The {\it propContent} denotes the content associated with the property. In the running example, ``every morning'' is the content associated with the property type {\it symprop/frequency}. 

Not all the symptoms mentioned in the course of clinical conversations are experienced by the patients. We explicitly infer the status of a symptom as experienced or not. This secondary task extracts the pair: ({\it symType, symStatus}).

\subsection{The Medication Task (Rx)}
This task consists of extracting tuples of the form: ({\it medContent, propType, propContent}). 

While symptoms can be categorized into a closed set, the set of medications is very large and continually updated. Moreover, in conversations, we would like to extract indirect references such as ``pain medications'' as {\it medContent}. We define three types of properties: medsprop/dosage, medsprop/duration and medsprop/frequency. In the running example,``twice a day'' is the {\it propContent} of the type {\it medsprop/frequency} associated with the {\it medContent} ``ibuprofen''. 

\section{Previous Work}
\label{sec:review}
Relation extraction is a long studied problem in the NLP domain and include tasks such as the ACE~\cite{Doddington2004-ix}, the SemEval~\cite{Hendrickx2010-dk}, the i2b2/VA Task ~\cite{Uzuner2011-ow}, and the BioNLP Shared Task~\cite{Kim2013-xx}. 
Many early algorithms such as DIPRE algorithm by \citet{Brin:1998} and SNOWBALL algorithm by \citet{Agichtein:2000} relied on regular expressions and rules~\cite{Fundel2007-nu, Peng2014-ra}. Subsequent work exploited syntactic dependencies of the input sentences. Features from the dependency parse tree were used in maximum entropy models~\cite{Kambhatla2004-ki} and neural network models~\cite{Snow:2005}. Kernels were defined over tree structures~\cite{Zelenko:2003,Culotta:2004,Qian2008-ae}. More efficient methods were investigated including shortest dependency path~\cite{Bunescu2005-bn} and sub-sequence kernels~\cite{Mooney:2005}. Recent work on deep learning models investigated convolutional neural networks~\cite{Liu:2013}, graph convolutional neural networks over pruned trees~\cite{Zhang:2018}, recursive matrix-vector projections~\cite{Socher:2012} and Long Short Term Memory (LSTM) networks~\cite{Miwa2016-bx}. Other more recent approaches include two-level reinforcement learning models~\cite{Takanobu_AAAI_2019}, two layers of attention-based capsule network models~\cite{Zhang_AAAI_2019}, and self-attention with transformers~\cite{N18-1080}. In particular,~\cite{miwa-sasaki-2014-modeling, katiyar-cardie-2016-investigating, zhang-etal-2017-end, zheng-etal-2017-joint, N18-1080, Takanobu_AAAI_2019} also seek to jointly learn the entities and relations among them together. A large fraction of the past work focused on relations within a single sentences. The dependency tree based approaches have been extended across sentences by linking the root notes of adjacent sentences~\cite{Gupta_AAAI_2019}. Coreference resolution is a similar task which requires finding all mentions of the same entity in the text ~\cite{Martschat2015-cl, Clark2016-uz, LeeHeLew17}.

In the medical domain, the BioNLP shared task deals with gene interactions and is very different from our domain~\cite{Kim2013-xx}. The {\it i2b2/va challenge} is closer to our domain of clinical notes, however, that task is defined on a small corpus of written discharge summaries~\cite{Uzuner:2011}. Written domain  benefits from cues such as the section headings which are unavailable in clinical conversations. For a wider survey of extracting clinical information from written clinical documents, see~\cite{Liu:2012}. 

 

\begin{figure*}[t]
\centering
\subfigure[Model architecture for the symptoms task.
\label{figure:sxtecture}]
{\includegraphics[width=0.49\linewidth]{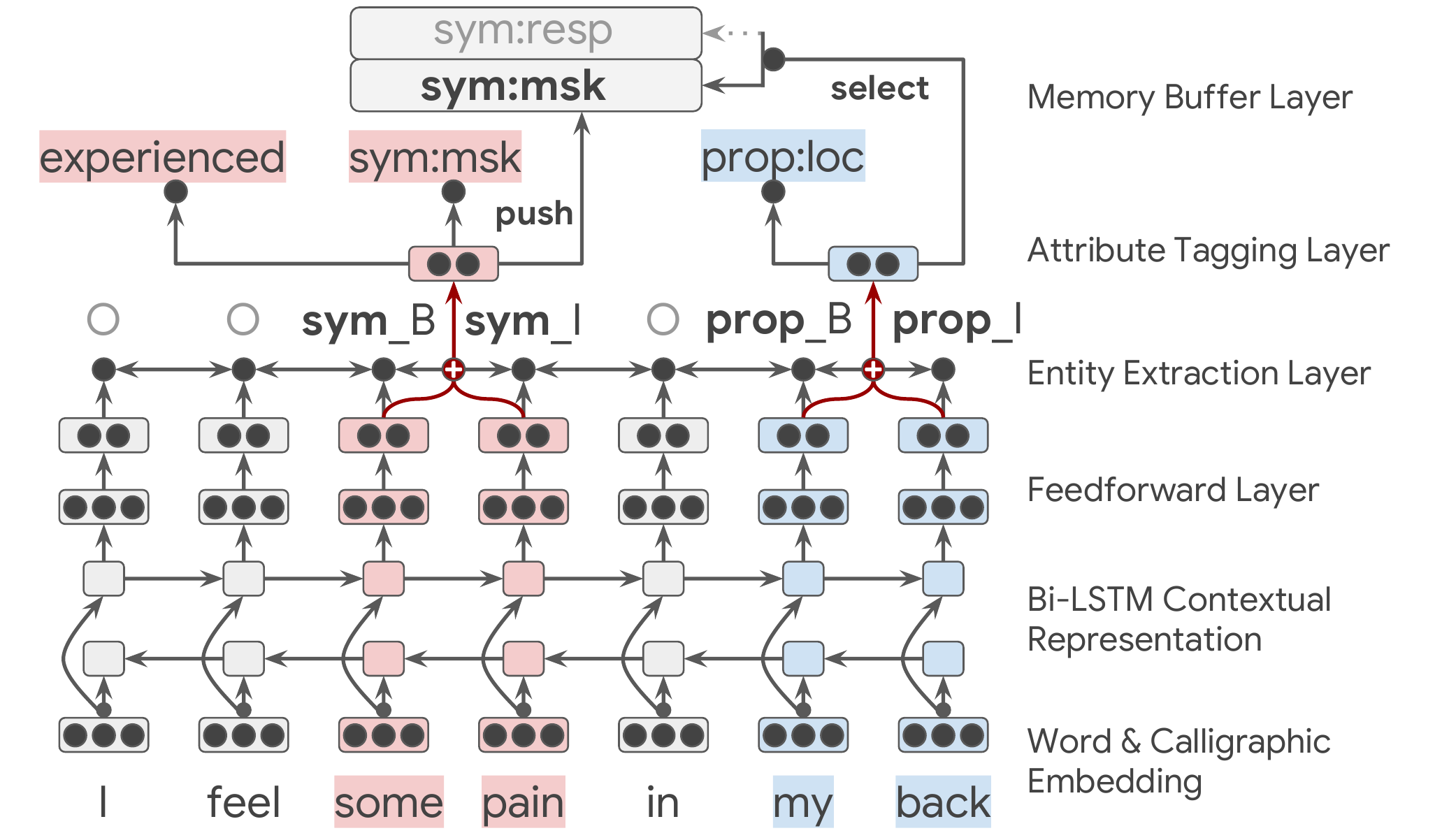}}
\subfigure[
Model architecture for the medications task.
\label{figure:rxtecture}
]{\includegraphics[width=0.49\linewidth]{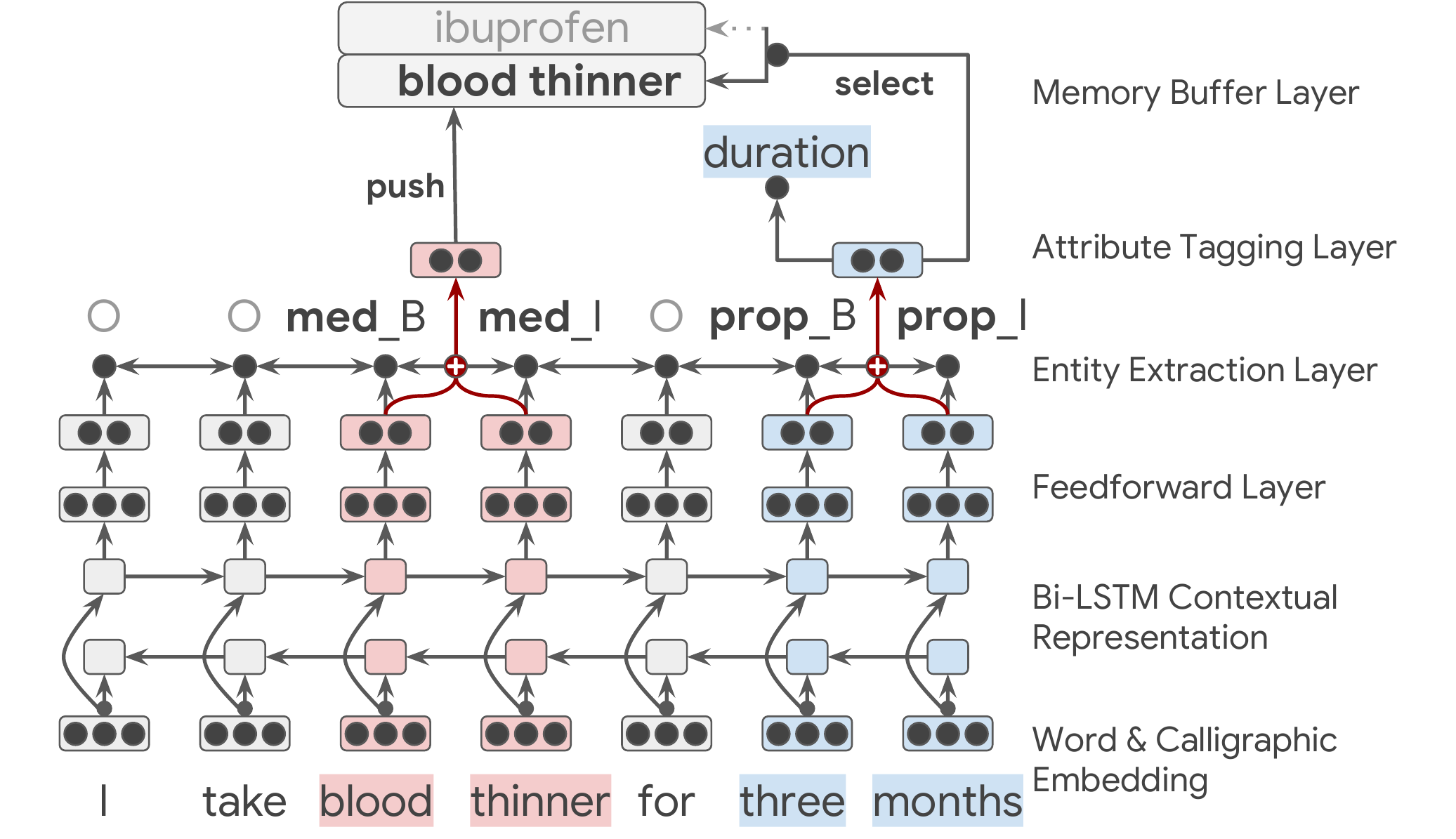}}
\caption{Variants of the R-SAT model architecture. The entity spans (``some pain'', ``blood thinner'') are identified in a tagging step, which are pushed into a memory buffer along with their latent representation and in a subsequent step the property spans (''my back'', ``three months'') selects the most related entity from the buffer.\label{rsat-architect}}
\end{figure*}

\section{Model} 

\label{sec:rsat}
Our application requires performing multiple inferences simultaneously, that of identifying symptoms, medications, their properties and relations between them. For this purpose, we adopt the well-suited multitask learning framework and develop a model architecture, illustrated in Figure~\ref{rsat-architect}, that utilizes our limited annotated corpus efficiently.

\subsection{Input Encoder Layer}\label{ssec:encoder}
Let $\mathbf{x}$ be an input sequence. We compute the contextual representation at the $k$-th step using a bidirectional LSTM, $\mathbf{h'_k}=[\vec{\mathbf{h}}(\mathbf{x_{\leq k}}|\vec{\Theta}_{LSTM}),\cev{\mathbf{h}}(\mathbf{x_{\geq k}}|\cev{\Theta}_{LSTM})]$, which is fed into a two-layer fully connected feed-forward network. For simplicity, we drop the index $k$ from the rest. The final features are represented as $\mathbf{h''} = MLP(\mathbf{h'}|\Theta_{FF})$. In our task, we found that the LSTM-based encoder performs better that the transformer-based encoder~\cite{transformer,mia}.

\paragraph{Extending a Standard Tagging Model} 
In a typical tagging model, the contextual representation of the encoder $\mathbf{h''}$ is fed into a conditional random field (CRF) layer to predict the BIO-style tags~\cite{ColWesBot11, HuaXuYu15, MaHov16, ChiNic16, LamBalSub16, PetAmmBha17, YanSalCoh17, ChaHuSha18}. Such a model can be extended to predict the relations. For example, in the utterance, ``I feel some \ul{pain} in \ul{my back}'', we could setup the tagger to predict the association between the symptom ({\it sym/msk}), and its property ({\it symprop/loc}) using a cross-product space where ``my back'' is tagged with {\it sym/msk+symprop/loc} so that the relation prediction problem is reformulated as a standard sequence labeling task. Although this would be a viable option for tasks where the tag set is small (e.g., place, organization, etc.), the cross-product space in our Sx task is unfortunately large (e.g., 186 Sx labels $\times$ 3 Sx property types, and 186 Sx labels $\times$ 3 Sx status types).

\subsection{Span Extraction Layer} \label{ssec:spanlayer}
We propose an alternative formulation that tackles the problem in a hierarchical manner. We first identify the span of interest using {\it\textbf{generic}} tags with BIO notation, namely, ({\it sym\_B}, {\it sym\_I}) for symptoms and ({\it symprop\_B}, {\it symprop\_I}) for their properties, as in Figure~\ref{figure:sxtecture}. Likewise, ({\it med\_B}, {\it med\_I}) for medications and ({\it medsprop\_B}, {\it medsprop\_I}) for their properties as shown in Figure~\ref{figure:rxtecture}. This corresponds to highlighting, for example, ``some pain'' and ``my back'' as spans of interest. 

Given the latent representations, $\mathbf{h}=(\bsy{h^{''}_1},\cdots,\bsy{h^{''}_N})$, and the target tag sequence $\mathbf{y}^e=(y_1, \cdots, y_N)$ (e.g., {\it{sym\_B}, {\it sym\_I}, {\it O}, {\it symprop\_B}, {\it symprop\_I}}), we use the negative log-likelihood $-\log P(\mb{y}^e|\mb{h})$ under CRF as the loss of identifying spans of interest $- S(\mathbf{y}^e, \mathbf{h}) + \log{\sum_{\mathbf{y}^\prime}\exp{(S(\mathbf{y}^\prime, \mathbf{h}))}}$, where $S(\mb{y}, \mb{h}) = \sum_{i=0}^N\mb{A}_{y_i, y_{i+1}} + \sum_{i=0}^N P(\mb{h}_i, \mb{y}_i)$ measures the compatibility between a sequence $\mb{y}$ and $\mb{h}$. The first component estimates the accumulated cost of transition between two neighboring tags using a learned transition matrix $\mb{A}$. $P(\mb{h}_i, \mb{y}_i)$ is computed via the inner product $\mb{h}_i^\top \mb{y}_i$ where $y_i$ belongs to any sequence of tags $\mb{y}$ that can be decoded from $\mb{h}$. During training, the $\log P(\mb{y}^e|\mb{h})$ is estimated using forward-backward algorithm and during inference, the most probable sequence is computed using the Viterbi algorithm. 

\subsection{Attribute Tagging Layer} \label{ssec:attributetagger}
Using the latent representation of the highlighted span, we can predict one or more {\it\textbf{attributes}} of the span. 
In Figure~\ref{figure:sxtecture}, we can predict two attributes associated with ``some pain'': {\it sym/msk} as the symptom label and {\it symStatus/experienced} as its status. Similarly, in Figure~\ref{figure:rxtecture}, the span property span ``three months'' has the predicted property type {\it medsprop/duration}. Therefore, by forming semantic abstractions for each highlighted text span, we decompose a single complex tagging task in a large label space into correlated but simpler sub-tasks, which are likely to generalize better when the training data is limited. 

Given the spans, either from the inferred or the ground truth sequence $\mb{y}^*$, a richer representation of the contexts can be used to predict attributes than otherwise possible. A contextual representation is computed from the starting $i$ and ending $j$ index of each span. 
\begin{align}
    \bsy{h}^s_{ij} = \AGGREGATE(\bsy{h}_k|\bsy{h}_k\in\mb{h}, i\leq k < j) \label{eqn:context}
\end{align}
where $\AGGREGATE(\cdot)$ is the pooling function, implemented as mean, sum or attention-weighted sum of the latent states of the input encoder. The $k$th attributes associated with the span are modeled using $P(y_{attr}^k|\bsy{h}^s_{ij})$. For example, while prediction symptom labels $s_x$ and their associated status $s_t$, the target attributes are $y_{attr}^0:=y^{s_x}$ and $y_{attr}^1:=y^{s_t}$. For predicting medication entities $r_x$ and their properties $p_r$, each span only has one attribute. Since each attribute comes from a pre-defined ontology, the multi-nomial distribution $P(y_{attr}^k|\bsy{h}^s_{ij})$ can be modeled as $\softmax(\bsy{h}^s_{ij}|\Theta^k)$ for each attribute. 

\subsection{Memory Buffer Layer} \label{ssec:buffer}
One of the critical components of our model is the memory buffer. Most previous models on joint inference of entities and relations consider all spans of entities and properties. This has the computational complexity of $\mathcal{O}(n^4)$ in the length of the input $n$, and makes it infeasible for application such as ours where the input could often be 1k words or more. We circumvent this problem using a memory buffer to cache all inferred candidate spans and test their relationship with inferred property spans. Note, unlike methods that cascade two such stages, our model is trained end-to-end jointly with multi-task learning.

The memory buffer saves different entries for symptom and medication tasks, as illustrated in Figure~\ref{rsat-architect}. At each occurrence of a symptom (medication) entity span, we push $\bsy{m}^k=\AGGREGATE(\{\bsy{h^{s}_{ij}}, \bsy{e^s}\})$ into the $k$-th position of the memory buffer. For the symptom task, $\bsy{e^s}$ is the learned word embedding of one of the labels in the closed label set. In the medication case, $\bsy{e^s}$ is the $\AGGREGATE$ of learned word embedding of the verbatim sub-sequence corresponding to the medication entity.

\subsection{Relation Inference Layer} \label{ssec:relation}
Each span of inferred property in the conversation is compared against each entry in the buffer. A property entity span is represented as $\bsy{y}^p=\AGGREGATE(\{\bsy{h^{p}_{ij}}, \bsy{e}^{p}\})$ where $\bsy{e}^p$ is the $\AGGREGATE$ of word embedding corresponding to the span. The multi-nomial likelihood is computed using a bilinear weight matrix $\bsy{W}$. The most likely entry ($k$) is picked from the memory stack $\bsy{M}=(\bsy{m}^1, ..., \bsy{m}^K)$ by maximizing the likelihood.
\begin{eqnarray}
\hat{k} &=& \argmax_k P(k|\bsy{y}^p) \nonumber \\
&=& \argmax_k \softmax({\bsy{y}^p}^\top \bsy{W} \bsy{M})
\end{eqnarray}

\paragraph{Remarks} The computation cost of inferring relation between a  property span and all the entities in the input is proportional to the memory buffer size. On our corpus, for Sx task, the mean and standard deviation per conversation was 22 and 15 respectively, and for Rx task, it was 32 and 23 respectively. Hence, the set of candidate entities considered is substantially smaller than all potential entities $O(n^2)$ in the input sequence. 

The small size of the memory buffer also has an impact on rate of learning. In each training step, rather than updating all embedding, we only update a smaller number of embedding, those associated with the entries in the memory buffer. This makes the learning fast and efficient.

\subsection{An End-to-end Learning Paradigm} \label{ssec:e2e_learning}
We train the model end-to-end by minimizing the following loss function for each conversation:
\begin{eqnarray}
{\cal L} = 
&-& \alpha\log P(\mb{y}^e|\mb{h}) \nonumber \\
&-& \sum_{\{y^{k}_{attr} \in S\}}\log P(y^{i}_{attr}|\mb{h}) \nonumber \\
&-& \sum_{\{y^{j}_{pos} \in P\}}\log P(y^{j}_{pos}|\mb{h}) 
\end{eqnarray}
where $\mb{y}^e$ is the target sequence $({\it sym\_B}$, ${\it sym\_I}$, ${\it prop\_B}$, ${\it prop\_I})$, $\{y^{k}_{attr}\}$ is the set of attribute labels for each highlighted span, $\{y^{j}_{pos}\}$ is the list of buffer slot indices and $\alpha$ is a relative weight. 

During training, we are simultaneously attempting to detect the location of tags as well as classify the tags. Initially our model for locating the tags is unlikely to be reliable, and so we adopt a curriculum learning paradigm. Specifically, we provide the classification stage the reference location of the tag from the training data with probability $p$, and the inferred location of the tag with probability $1-p$. We start the joint multi-task training by setting this probability to $1$ and decrease it as training progresses~\cite{BenVinJaiSha15}.

Since our model consists of span extraction and attribute tagging layers followed by relation extraction, we refer to our model as {\bf Relational Span-Attribute Tagging Model} (R-SAT). One advantage of our model is that the computational complexity of joint inference is  $\mathcal{O}(n)$ which is linear in the length of the conversation $n$. This is substantially cheaper than other previous work on joint relation prediction models where the computational complexity is $\mathcal{O}(N^4)$~\cite{LeeHeLew17}.

\section{Knowledge Graph Features} \label{ssec:kg}
Medical domain knowledge could be helpful in increasing the likelihood of symptoms when related medications is mentioned in a conversation, and vice versa. One such source is a knowledge graph (KG) whose embedding represent a low-dimensional projection that captures structural and semantic information of its nodes. Previous work has demonstrated that KG embedding can improve relation extraction in written domain~\cite{Han2018-np}. 
We utilize an internally developed KG that contains about 14k medical nodes of 87 different types (e.g., medications, symptoms, treatments, etc.).
\begin{figure}
\centering
\includegraphics[width=0.85\linewidth]{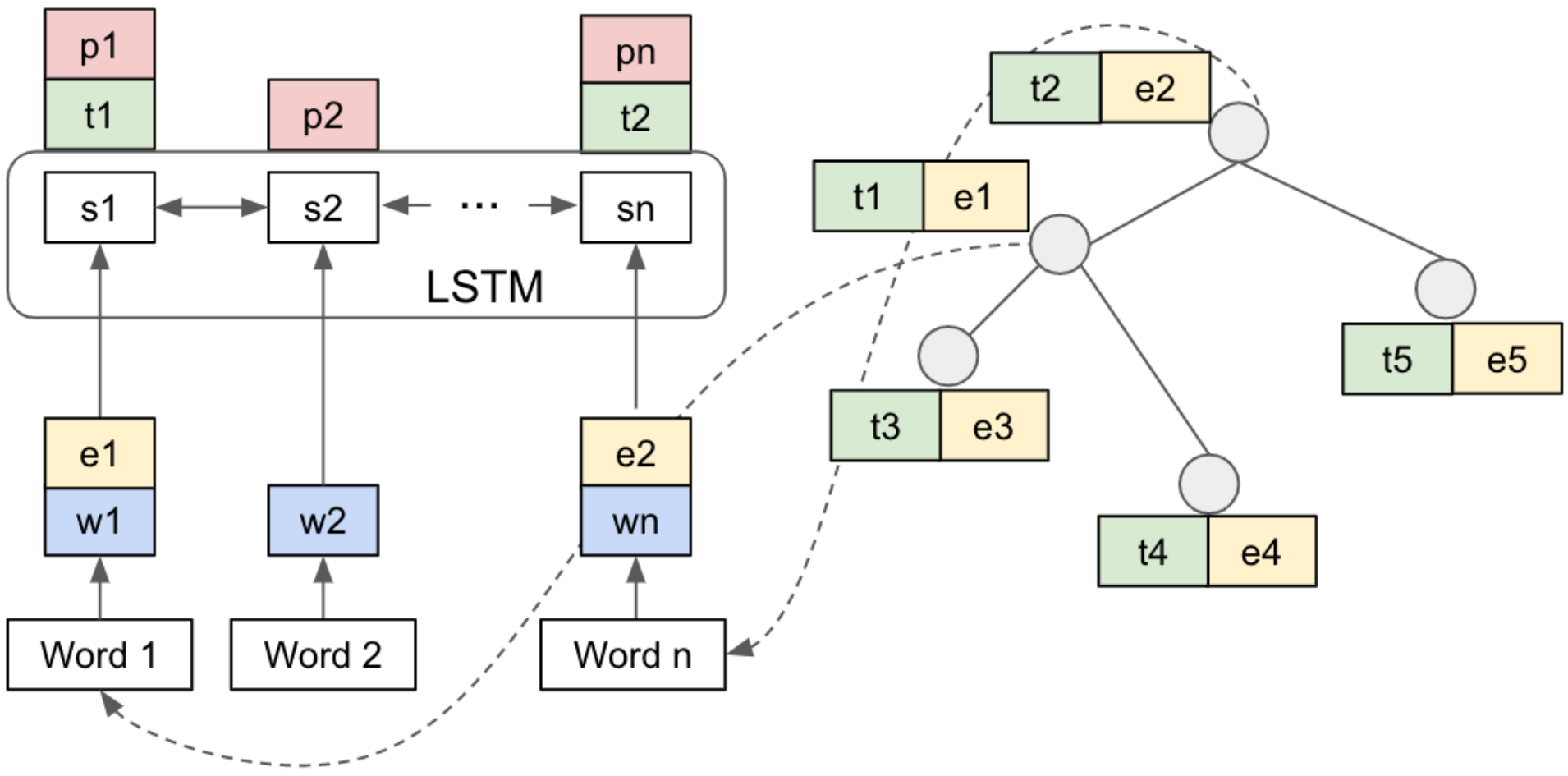} 
\caption{Illustration of how POS ($p$) features and knowledge graph (KG) are incorporated into our encoder. Dashed lines represent mappings from words ($w$) to KG nodes, which contain the embedding ($e$) and the type ($t$) information.}
\label{figure:kg} 
\vspace{-4mm}
\end{figure}
The nodes are represented by 256 dimension embedding vectors, which were trained to minimize word2vec loss function on web documents~\cite{TomIlyKaiGreJef13}. A given node may belong to multiple types and this is encoded as a sum of one-hot vectors. The input word sequences were mapped to KG nodes using an internal tool ~\cite{Brown2013globally}. For words that do not map to KG nodes, we use a learnable UNK vector of the same dimension as the KG embedding. In addition, we also represented linguistic information using part-of-speech (POS) tags as one-hot vector. The POS tags were inferred from the input sequence using an internal tool with 47 distinct tags~\cite{Andor2016globally}.
In our experiments, we find it most effective to concatenate word embedding with KG entities, and the encoder output with the embedding of POS tags and the KG entity types. 

\section{Experiments} \label{sec:expts}
We describe our corpus, evaluation metrics, the experimental setup, the evaluations of the proposed model and comparison with different baselines on both the symptom and medication tasks.

\subsection{Corpus} \label{ssec:corpus}
Given the privacy-sensitive nature of clinical conversations, there aren't any publicly available corpora for this domain. Therefore, we utilize a private corpus consisting of 92K de-identified and manually transcribed audio recordings of clinical conversations, typically about 10 minutes long [IQR: 5-12 minutes] with mostly 2 participants (72.7\%). Other participants when present included, for example, nurses and caregivers. The corresponding manual transcripts contained sequences that were on average 208 utterances or 1,459 words in length. We note that due to the casual conversational style of speech, an entity mentioned at the beginning can be related to a property mentioned at the end of the conversation. This makes the problem of modeling relations much harder than previous work on extracting relations.

A subset of about 2,950 clinical conversations, related to primary care, were annotated by professional medical scribes. The ontology for labeling medication consisted of the type of medications (e.g., medications, supplements) and their properties (e.g., dosage, frequency, duration), and that for symptoms consisted of 186 symptom names and their properties. This resulted in 77K and 99K tags for the medication and symptom tasks, respectively. In all, there were 23k and 16k relationships between medications and symptoms and their properties, respectively. The conversations were divided into training (1,950), development (500) and test (500) sets.

In the case of medications, about 70\% of the labels were about medications and the rest about their properties, of which 51\% were dosage or quantity, and 40\% were frequency. In the case of symptoms, 41\% of the labels were about symptom names, another 41\% about status, and the rest about properties, of which 39\% were about frequency and 37\% about body locations. 

\subsection{Pretraining} \label{ssec:pretraining}
Since our labeled data is small, only about 3k, the input encoder of the model was pre-trained over the entire 92k conversations. For pre-training, given a short snippet of conversation, the model was tasked with predicting the next turn, similar to skip-thought~\cite{Ryan_2015}. Our models were trained using the Adam optimizer~\cite{KinBa15} and the hyperparameters are described in the supplementary material.

\subsection{Evaluation Metrics} \label{ssec:metrics}
As described in Section~\ref{sec:task}, our tasks consist of extracting tuples -- ({\it symType, propType, propContent}) for symptoms task and ({\it medContent, propType, propContent}) for medications. The precision, recall and F1-scores are computed jointly over all the three elements and the content is treated as a list of tokens for evaluation purposes. To allow for partial content matches, we generalize the calculation of precision and recall such that
\begin{align*}
    Precision = \frac{1}{|\mathcal{S}_{\hat{Y}}|}\sum_{i \in \mathcal{S}_{\hat{Y}}} \prod_{j=1}^{3} &\mathcal{I}_{\hat{y}_i^j}(\hat{y}_{i}^{j}, y_{i}^{j}) \\
    Recall = \frac{1}{|\mathcal{S}_{Y}|}\sum_{i \in \mathcal{S}_{Y}} \prod_{j=1}^{3} &\mathcal{I}_{y_i^j}(\hat{y}_{i}^{j}, y_{i}^{j})
\end{align*}
\noindent
where $\mathcal{S}_{\hat{Y}}$ denotes the set of predictions, $\mathcal{S}_{Y}$ denotes the set of ground truths, and $\mathcal{I}_{z_i^j}(\hat{x}_i^{j}, x_i^{j}) = |\hat{x}_i^{j} \cap x_i^{j}| / |z_i^{j}|$. We note that, as \textit{symType} and \textit{propType} are simply target classes, $\mathcal{I}$ reduces to a simple indicator function. Under the scenario that the content includes single elements, the entire calculation simplifies to the exact matching-based calculation of precision and recall over the set of predictions and ground truths. For the symptom task, we additionally evaluate the performance of predicting \textit{symType} and \textit{symStatus} by performing the exact matching-based calculation. 

We illustrate this evaluation metric with an example below: 
\begin{table}[ht]
    \centering
    \begin{tabular}{l}
Prediction: [(sym/sob, prop/severity, [bad])] \\
Reference: [(sym/unk, prop/location, [arm]), \\
\hspace{.6in} (sym/sob, prop/severity, [really, bad])] \\
    \end{tabular}
\end{table} 
There are two symptoms in the reference and the model extracted one of them. In the extracted symptom, the model correctly identified one out of the two content words. So, we score the precision as $1/1(1*1*(1/1))=1$ and recall as $1/2((0*0*0) + (1*1*(1/2))) = 0.25$.

\subsection{Baselines} \label{ssec:baseline}

\paragraph{Symptom Task} As a baseline for this task, we train an extension of the standard tagging model, described in Section~\ref{ssec:encoder}. The label space for extracting the relations between symptoms and their properties is {\it 186 symptoms $\times$ 3 properties}, and for extracting symptoms and their status is {\it 186 symptoms $\times$ 3 status}. Using the BIO-scheme, that adds up to 2,233 labels in total. The baseline consists of a bidirectional LSTM-encoder followed by two feed-forward layers $[512, 256]$ and then a 2,233 dimension softmax layer. The label space is too large to include a CRF layer. The encoder was pre-trained in the same way as described in Section~\ref{ssec:pretraining}, the hyperparameters were selected according to Table~\ref{table:ParamSeq2Seq}, and the model parameters were trained using cross-entropy loss.

\paragraph{Medication Task} For this task, we adopt a different baseline since the generic medication entity type (e.g., drug name, supplement name) does not provide any useful information unlike the 186 symptom entity labels (e.g., sym/msk/pain). Instead, we adopt the neural co-reference resolution approach which is better suited to this task~\cite{LeeHeLew17}. The encoder is the same as the baseline for symptom task and pre-trained in the same manner. Since the BIO labels contain only 9 elements in this case, the encoder output is fed into a CRF layer. Each candidate relation is represented by concatenating the latent states of the head tokens of the medication entity and the property. This representation is augmented with an embedding of the token-distance, which is fed to a softmax layer whose binary output encodes whether they are related or not. Note our R-SAT model does not take the advantage of this distance embedding.

\subsection{Parameter Tuning}
\label{sec:hyperparam}
Table~\ref{table:ParamSeq2Seq} shows the parameters that were selected after evaluating over a range on a development set. In all experiments, the $\AGGREGATE(\cdot)$ function is implemented as the mean function for its simplicity.
\begin{table}[H]
\small
\renewcommand\tabcolsep{4pt}
  \centering
  \begin{tabular}{llll}
  \hline
  Parameter & Used  & Range \\ \hline
  Word emb & 256  & [128 -- 512] \\
  LSTM Cell & 1024  & [256 -- 1024] \\
  Enc/dec layers & 1  & [1 -- 3] \\
  Dropout & 0.4 & [0.0 -- 0.5] \\
  L2 & 1e-4 & [1e-5 -- 1e-2]\\
  Std of VN & 1e-3 & [1e-4 -- 0.2] \\
  $\alpha$ & 0.01 & [1e-4 -- 0.1]\\
  Learning rate & 1e-2 & [1e-4 -- 1e-1]\\
  \hline
  \end{tabular}
  \caption{Hyperparameters of our models for model reproducibility.}
  \label{table:ParamSeq2Seq}
  \vspace{-2mm}
\end{table}

\subsection{Results \& Ablation Analysis} \label{ssec:ablation}
The performance of the proposed R-SAT model was compared with the baseline models, and the results are reported in Table~\ref{table:ablation}.
\paragraph{Symptom Task} The model was trained using multi-task learning for both tasks: ({\it symType, propType, propContent}) as well as ({\it symType, symStatus}). The performance was evaluated using all the elements of the tuple as described in Section~\ref{ssec:metrics}. The baseline performs better on ({\it symType, symStatus}) compared to ({\it symType, propType, propContent}) possibly because there are more instances of the former in the training data than the latter. The R-SAT model performs significantly better than baselines on both tasks. 

\begin{table*}[ht]
  \centering
  \begin{tabular}{l|cc|c}
  \hline
  Model                                         & Sx + Property & Sx + Status & Rx + Property \\ \hline
  Baseline                                             & 0.18       & 0.44       & 0.35 \\
  R-SAT                                                & {\bf 0.34} & {\bf 0.57} & {\bf 0.45} \\
  \hspace{3pt} w/o [KG]                                & 0.30       & 0.56       & 0.43 \\
  \hspace{3pt} w/o [KG, Context]                       & 0.26       & 0.55       & 0.30 \\
  \hspace{3pt} w/o [KG, Context, Buffer]               & 0.24       & 0.55       & n/a  \\
  \hspace{3pt} w/o [KG, Context, Buffer, Multi-task]   & 0.23       & n/a        & n/a  \\
  \hline
  Human                                                & 0.51       & 0.78       & 0.52 \\
  \hline
\end{tabular}
\caption{Comparison of the performance of the proposed R-SAT model with baselines and ablation analysis on different components (KG, Context, Buffer, Multi-task) where `context' is the latent representation of the span $\bsy{h^{s}_{ij}}$ in the memory buffer.}
\label{table:ablation}
\end{table*}

For understanding the contribution of different components of the model, we performed a series of ablation analysis by removing them one at a time. In extracting relations in {\it Sx + Property}, the KG embeddings along with POS tags contribute a relative gain of {\bf 13\%} while the memory buffer brings a relative gain of {\bf 8\%}. Neither of them impact {\it Sx + status}, and that is expected for memory buffer since the status is tagged on the same span as the contents of the memory buffer. Multi-task learning brings a relative improvement of {\bf 4\%} on {\it Sx + Property}, and this may be because there are fewer instances of this relation in the training data, and jointly learning with {\it Sx + Status} helps to learn better representations. Note we have not checked other sequences for removing model components (e.g., removing Multi-tasking earlier or KG later).

\paragraph{Medication Task} In the Rx case, we only have one task ({\it Rx + Property}), that is, predicting the relations between medications and their properties, e.g., ([ibuprofen], prop/dosage, [10 mg]). The baseline gives reasonable performance. Ablation analysis reveals that KG and POS features contribute about {\bf 4.6\%} relative improvement, while the contextual span in memory buffer adds a substantial {\bf 43\%} relative improvement. Since the medications are from an open set, we cannot run experiments without the buffer. Compared to symptoms task, the model performs better on medication task, and this may be due to lower variability in dosage.

\paragraph{Relation Only Prediction} For teasing apart the strength and weakness of the model, we evaluated its performance when the entities and their properties were given, and the model was only required to decide whether a relation exists or not. 

\begin{table}
  \centering
  \begin{tabular}{l|c|c}
  \hline
  Model & {\it Sx + Property} & {\it Rx + Property} \\ \hline
  $\text{BRAN}$ & 0.62 & 0.41 \\
  \hline
  R-SAT & 0.82 & 0.60 \\
  \hline
\end{tabular}
\caption{Performance of the model when the entities and properties are given and it is only required to predict existence of relations.}
\label{table:gold-result}
\end{table}

As a baseline, we compare our model with a most recently proposed model for document-level joint entity and relation extraction: BRAN, which achieved state-of-art performance for chemical-disease relation~\cite{N18-1080}. 
When this model was originally used to test relations between all pairs of entities and properties in the entire conversation, it performed relatively poorly. Using the implementation released by the authors, the performance of BRAN was then optimized by restricting the distance between the pairs and by fine-tuning the threshold. The best results are reported in Table~\ref{table:gold-result}. Our proposed R-SAT model without any such constraints performs better than BRAN on both tasks by an absolute F1-score gain of about 0.20.

Interestingly, the performance of our model on {\it Sx + Property} jumps from 0.34 in the joint prediction task to 0.82 in the relation only prediction task. This reveals the primary weakness of the Sx model is in tagging the entities and the properties accurately. In contrast, the F1-score for {\it Rx + Property} is impacted less, and only moves up from 0.45 to 0.6.

The task of inferring whether a relation is present between a medication and its properties is more challenging than in the case of symptoms task. This is not entirely surprising since there is a higher correlation between symptom type and location (e.g., respiratory symptom being associated with nose) and relatively low correlation between dosage and medications (e.g., 400mg could be the dosage for several different medications).

\subsection{Analysis} \label{sec:analysis}
For understanding the inherent difficulty of extracting symptoms and medications and their properties from clinical conversations, we estimated human performance. A set of 500 conversations were annotated by 3 different scribes. We created a ``voted'' reference and compared the 3 annotations from each of the 3 scribes against them. 

The F1-score of scribes were surprisingly low, with 0.51 for {\it Sx + Property} and 0.78 for {\it Sx + Status}. The model performance also finds extracting relation in {\it Sx + Property} to be more difficult than {\it Sx + Status} task. In summary, the model performance reaches 67\% of human performance for {\it Sx + Property} and 73\% for {\it Sx + Status}. The F1-score of scribes for {\it Rx + Property} is similar to that of {\it Sx + Property}. In this case, the model achieves about 85\% of human performance. The human errors or inconsistencies in Sx and Rx annotations appear to be largely due to missed labels and not due to inconsistent spans for the same tags, or inconsistent tags for the same span.

While the majority of our relations in the reference annotations occurred within the same sentence, approximately 11.1\% of relations occurred across 3 or more sentences. This typically occurred when the symptoms or medications are discussed over multiple dialog turns, as illustrated in Table \ref{table:task}. Among the relations correctly identified by the model, 10.6\% were also across 3 or more sentences, which is very similar to the priors on the reference and seem to contain no bias. We notice that in certain cases, the model is able to link a property to an entity that is far away (100+ sentences) when a nearby mention of the same entity was missed by the model. Models that only examine relations in nearby sentences (2-3 sentences) would have missed the relation in such a scenario.

The majority of the errors result from our model missing the property span. Specifically, we see that 35\% and 81\% of the errors are due to model not detecting medications and symptoms property. For example, “when i really have to”, “every three three months”, which are rare mentions in informal language. 

Our reference links each property to only one entity. In certain cases, we notice that the model links the entity to an alternative mention or entity that is equally valid (Advil vs pain killer). So, our performance measure underestimates the actual model performance.

\section{Conclusions} \label{sec:conclusions}
We propose a novel model to jointly infer entities and relations. The key components of the model are: a mechanism to highlight the spans of interest, classify them into entities, store the entities of interest in a memory buffer along with the latent representation of the context, and then infer relation between candidate property spans with the entities in the buffer. The components of the model are not tied to any domain. We have demonstrated applications in two different tasks. In the case of symptoms, the entities are categorized into 188 classes, while in the case of medications, the entities are an open set. The model is tailored for tasks where the training data is limited and the label space is large but can be partitioned into subsets. The two stage processing where the candidates are stored in a memory buffer allows us to perform the joint inference at a computational cost of ${\cal O}(n)$ in the length of the input $n$ compared to methods that explore all spans of entities and properties at a computational cost of ${\cal O}(n^4)$. The model is trained end-to-end. We evaluate the performance on three related tasks, namely, extracting symptoms and their status, relations between symptoms and their properties, and relations between medications and their properties. Our model outperforms the baselines substantially, by about 32-50\%. Through ablation analysis, we observe that the memory buffer and the KG features contribute significantly to this performance gain. By comparing human scribes against ``voted'' reference, we see that the task is inherently difficult, and the models achieve about 67-85\% of human performance.  

\section*{Acknowledgments}
This work would not have been possible without the help of a number of colleagues, including Laurent El Shafey, Hagen Soltau, Yuhui Chen, Ashley Robson Domin, Lauren Keyes, Rayman Huang, Justin Stuart Paul, Mark Knichel, Jeff Carlson, Zoe Kendall, Mayank Mohta, Roberto Santana, Katherine Chou, Chris Co, Claire Cui, and Kyle Scholz.

\bibliography{medical_relations}
\bibliographystyle{acl_natbib}

\end{document}